\documentclass[10pt,twocolumn,letterpaper]{article}

\usepackage{iccv}
\usepackage{times}
\usepackage{epsfig}
\usepackage{graphicx}
\usepackage{amsmath}
\usepackage{amssymb}

\usepackage{bm}
\usepackage{booktabs}
\usepackage{multirow}
\usepackage[dvipsnames]{xcolor}
\usepackage{color}
\usepackage[pagebackref=true,breaklinks=true,colorlinks,bookmarks=false,citecolor=blue,linkcolor=blue]{hyperref}
\usepackage{arydshln} 
\usepackage{bbding}
\usepackage[breaklinks=true,bookmarks=false]{hyperref}

\iccvfinalcopy 

\def\iccvPaperID{4560} 

\ificcvfinal\fi

\begin{document}

\title{Sign Language Translation with Iterative Prototype}

\author{Huijie~Yao$^{1}$~~~~~Wengang~Zhou$^{1,2,}$\thanks{Corresponding authors: Wengang Zhou and Houqiang~Li}~~~~~Hao~Feng$^{1}$~~~~~Hezhen~Hu$^{1}$~~~~~Hao~Zhou$^{1}$~~~~~Houqiang~Li$^{1,2,}$\footnotemark[1] \\
{\normalsize $^{1}$ CAS Key Laboratory of Technology in GIPAS, EEIS Department, University of Science and Technology of China} \\
{\normalsize $^{2}$ Institute of Artificial Intelligence, Hefei Comprehensive National Science Center} \\
{\tt\small \{yaohuijie,haof,alexhu,zhouh156\}@mail.ustc.edu.cn, \{zhwg,lihq\}@ustc.edu.cn}
}

\maketitle

\ificcvfinal\thispagestyle{empty}\fi

\begin{abstract}
This paper presents IP-SLT, a simple yet effective framework for sign language translation (SLT). Our IP-SLT adopts a recurrent structure and enhances the semantic representation (prototype) of the input sign language video via an iterative refinement manner. Our idea mimics the behavior of human reading, where a sentence can be digested repeatedly, till reaching accurate understanding. Technically, IP-SLT consists of feature extraction, prototype initialization, and iterative prototype refinement. The initialization module generates the initial prototype based on the visual feature extracted by the feature extraction module. Then, the iterative refinement module leverages the cross-attention mechanism to polish the previous prototype by aggregating it with the original video feature. Through repeated refinement, the prototype finally converges to a more stable and accurate state, leading to a fluent and appropriate translation. In addition, to leverage the sequential dependence of prototypes, we further propose an iterative distillation loss to compress the knowledge of the final iteration into previous ones. As the autoregressive decoding process is executed only once in inference, our IP-SLT is ready to improve various SLT systems with acceptable overhead. Extensive experiments are conducted on public benchmarks to demonstrate the effectiveness of the IP-SLT.
\end{abstract}

\section{Introduction}
\begin{figure}
\centering
\includegraphics[width=80mm]{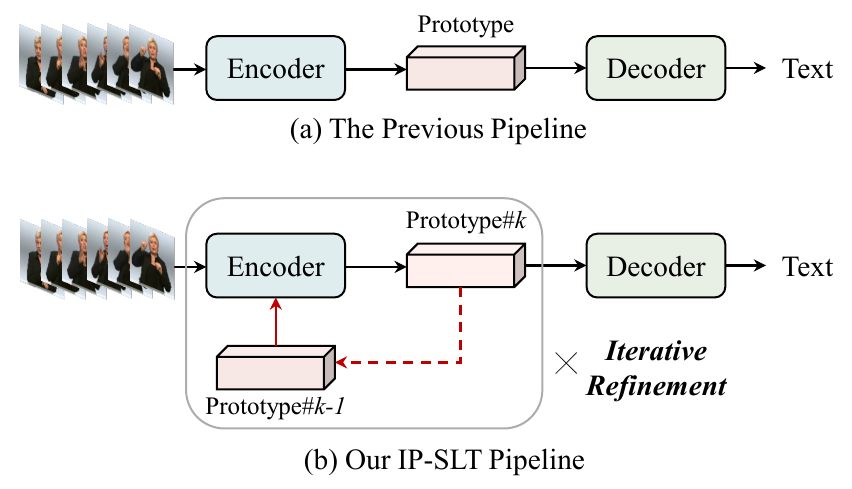}
\caption{Illustration of the pipeline of the previous works and our IP-SLT. (a) The previous studies rely on a one-pass forward process to generate the final translation. (b) To mitigate the vision-text gap, we introduce the iterative refinement module into the original SLT system. The refinement module updates the current prototype conditioned on the sign language video, which can be run iteratively to obtain a better representation of the semantic meaning of the sign language video.
}
\label{fig1}
\end{figure}

Sign language translation (SLT) aims to automatically generate spoken language translations based on sign language videos, which holds both social significance and academic value. On the one hand, a high-quality SLT system can greatly facilitate communication between deaf-mute and hearing individuals~\cite{bragg2019sign,cooper2011sign,cormier2019extol,sutton1999linguistics}. On the other hand, SLT as an interdisciplinary research topic necessitates a comprehensive understanding of computer vision~\cite{min2021visual, cui2017recurrent, pu2019iterative,he2022revisiting} and natural language processing~\cite{vaswani2017attention, sutskever2014sequence}, given its involvement with vision and text modalities. As a result, SLT has emerged as a vital research topic, garnering increasing attention~\cite{camgoz2018neural,zhou2021improving,min2021visual,albanie2020bsl,huang2018video,hu2023signbert+,tang2021graph}.

SLT is a challenging task, which faces a tough domain gap between the input video and output text, as well as a limited dataset scale due to costly data collection and annotation~\cite{zhou2021improving,camgoz2020sign,wang2016isolated, huang2018video}. Since SLT is typically viewed as a sequence-to-sequence mapping problem, the existing SLT systems~\cite{jin2022prior,camgoz2018neural,camgoz2020sign} commonly adopt the one-pass forward pipeline based on encoder-decoder architecture~\cite{sutskever2014sequence,luong2015effective} (as shown in Fig.~\ref{fig1} (a)). 
In such a framework, the encoder transforms the sign video into its semantic representation (prototype), which is then fed into the decoder to obtain the final translation. 
However, due to the inherent gap between vision and text, it may be hard to conduct such mapping within the vanilla one-pass architecture.

In this study, we present IP-SLT with the iterative prototype to boost sign language translation (as shown in Fig.~\ref{fig1} (b)), which is inspired by the human reading process.
During this process, we note that repeatedly digging into the source materials is necessary for accurate understanding.
Similarly, when we are trying to translate a sign language video into a sentence, we commonly do not directly write it down. Instead, we would recall and go back to the original sign video to check our answers. To implement the above idea, our IP-SLT adopts a recurrent structure that enhances the semantic representation (prototype) of the input sign language video via an iterative refinement process. 
IP-SLT generally contains three main components, including feature extraction, prototype initialization, and iterative prototype refinement. 
Given a sign video to be translated, we first extract its visual representation, which is then used 
for generating an initial raw prototype. 
Subsequently, we iteratively leverage the attention mechanism~\cite{vaswani2017attention} to update the prototype toward the semantic meaning of the sign video.
At each iteration, we refine the previous prototype by aggregating it with the original visual representation.
In this way,
the network repeatedly digs the semantic context from the sign video to polish the prototype.
Through iterative refinement, the prototype finally converges to a stable and accurate state, producing a high-quality translation. 

In addition, 
our IP-SLT introduces a novel design discussed next.
Firstly, to leverage the sequential dependence between different iterations, we further propose the iterative distillation loss which allows the previous prototypes to obtain supervision from the final one. Since the final prototype converges to a more stable and accurate state, it is possible for IP-SLT to achieve better performance.
Secondly, during training, all predicted prototypes are transformed into their corresponding translations to provide guidance for each iteration.
Our inference process is neat since only the final prototype is used for the autoregressive decoding process.
Thirdly, our IP-SLT can easily work with different visual backbones. 
Through end-to-end optimization, our IP-SLT achieves significant performance improvements over the baselines.

In summary, our contributions are three-fold:
\begin{itemize}
\item[$\bullet$] We propose IP-SLT, a novel framework to ameliorate sign language translation, which iteratively refines the prototypes by aggregating the previous translation progress and the original visual representation.
\end{itemize}
\begin{itemize}
\item[$\bullet$] We propose an iterative distillation loss to enhance the basic supervision, by leveraging the sequential dependence between the outputs at each iteration.
\end{itemize}
\begin{itemize}
\item[$\bullet$] 
We conduct extensive experiments to validate the proposed method, and show encouraging improved results on the two prevalent benchmarks, \emph{i.e.}, CSL-Daily~\cite{zhou2021improving} and PHOENIX-2014T~\cite{camgoz2018neural}.
\end{itemize}

\section{Related Work}

\begin{figure*}
\centering
\includegraphics[width=150mm]{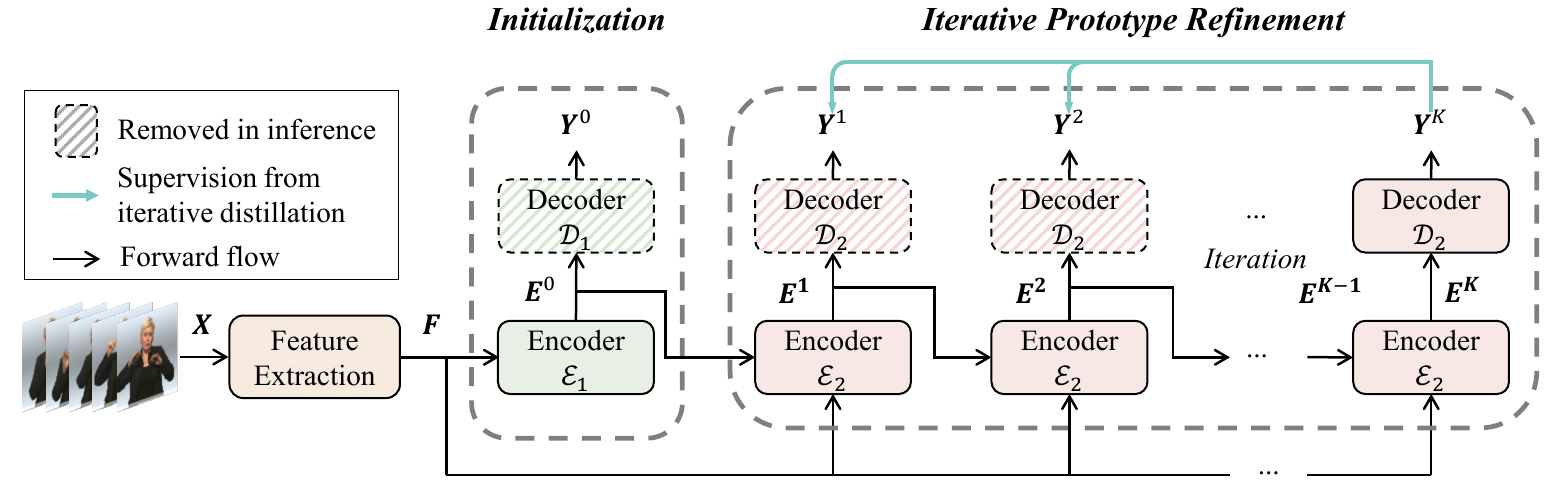}
\caption{An overview of the proposed IP-SLT framework. Given a sign video $\bm{X}$, the feature extraction module is responsible for embedding the input into visual representation $\bm{F}$. The initialization module (the encoder $\mathcal{E}_1$ and decoder $\mathcal{D}_1$) generates the initial prototype $\bm{E}^0$ and the raw translation $\bm{Y}^0$. The refinement module (the encoder $\mathcal{E}_2$ and decoder $\mathcal{D}_2$) first takes the initial prototype $\bm{E}^0$ as input and generates a prototype for the current step by fusing it with the original visual representation $\bm{F}$. Through $K$ times refinement, the prototype sequence $\bm{E}=\{\bm{E}^0,\bm{E}^1,\cdots,\bm{E}^K\}$ and corresponding translation sequence $\bm{Y}=\{\bm{Y}^0,\bm{Y}^1,\cdots,\bm{Y}^K\}$ are obtained.
In light of the fact that the decoding part of IP-SLT consists of $K+1$ branches based on the iteration order, we introduce the iterative distillation loss to improve the underlying supervision. It should be noted that the parts enclosed in dashed boxes can be removed in inference.
}
\label{fig2}
\end{figure*}
In this section, we briefly review the related works, \emph{i.e.}, sign language translation and iterative refinement methods.

\smallskip
\noindent \textbf{Sign Language Translation}. 
Camgoz~\etal.~\cite{camgoz2018neural} pioneer the neural SLT task and publish the neural dataset PHOENIX-2014T and regard the SLT as a sequence-to-sequence problem.
They implement the neural SLT system using the encoder-decoder paradigm~\cite{bahdanau2014neural}. 
This paradigm is adopted by subsequent studies which focus on addressing the challenges of data scarcity and domain gap. 
Considering the lack of frame-level annotation in sign language datasets, Li~\etal~\cite{li2020tspnet} design the temporal semantic pyramid structure to obtain more discriminative features. 
Camgoz~\etal~\cite{camgoz2020sign} explore the mutual benefits of SLT and continuous sign language recognition through joint optimization. Zhou ~\etal~\cite{zhou2021improving} leverage gloss annotation to transform the monolingual texts into pseudo-videos. According to the characteristic of sign language, several works~\cite{yin2020better,camgoz2020multi,zhou2021spatial} propose multi-channel SLT systems which explicitly extract and align the key parts of sign language expression. Jin~\etal~\cite{jin2022prior} leverage the additional prior knowledge to obtain high-quality translations. 
Chen~\etal~\cite{chen2022simple} propose a transfer learning baseline for SLT by leveraging external resources from related tasks.
Chen~\etal~\cite{chen2022two} further combine the raw videos and the keypoint sequences to achieve better semantic understanding with auxiliary supervision.

In contrast, our proposed approach employs an iterative refinement process that utilizes the previous prototype as an additional clue to enhance the accuracy of the mapping between sign videos and their translations.

\smallskip
\noindent \textbf{Iterative Refinement Methods}.
The idea of iterative refinement is applied to various computer vision tasks, such as image generation~\cite{ahn2018image,gregor2015draw,ren2019progressive}, instance segmentation~\cite{peng2020deep,ling2019fast,liu2021dance,zhang2022e2ec}, image rectification~\cite{feng2021docscanner}, \etc., which shows promising performance improvements. CARN~\cite{ahn2018image} maintains training stability in super-resolution tasks and improves the quality of output images. Ling~\etal~\cite{ling2019fast} regard the object instance segmentation task as a regression task and proposes the Curve-GCN to iteratively predict the locations of all vertices. DeepSanke~\cite{peng2020deep} uses the deep network to iteratively enclose the object boundary based on an initial contour.
The previous studies~\cite{zhou2019dynamic,pu2018dilated,cui2019deep,pu2019iterative,min2021visual} in continuous sign language recognition  task follow an iterative training scheme to enhance the discriminative power of feature extraction modules which use the convolutional neural network and their variants~\cite{carreira2017quo, ji20123d, simonyan2014two}. They leverage alignment proposals given by the connectionist temporal classification (CTC)~\cite{graves2006connectionist} decoding as supervision at frame-wise granularity, which cannot be directly applied to the SLT. 

Different from the aforementioned methods, we explore how to reduce the vision-text gap by proposing an iterative refinement module to the existing SLT system. To reduce the complexity, we design the refinement module in a shared-weight manner.
Moreover, we put forward the iterative distillation loss to leverage the sequential dependence between different iterations.

\section{Methodology}

In this section, we first introduce the overall architecture of our IP-SLT, and then separately elaborate individual components. Finally, we propose the design of the training objective and inference strategy for the IP-SLT.

    \subsection{Framework Overview}
    The primary objective of the SLT system is to acquire knowledge about the mapping $f: \mathcal{X} \mapsto \mathcal{Y}$, where $\mathcal{X}$ and $\mathcal{Y}$ denote the collections of $N$ sign language videos and spoken language sentences associated with vocabulary $\mathcal{V}$, respectively. Most SLT systems adopt the encoder-decoder architecture~\cite{sutskever2014sequence}, where the input $\bm{X} \in \mathcal{X}$ is first encoded to derive a high-level context representation. It is then passed to the decoder to generate the output $\bm{Y} \in \mathcal{Y}$. The encoder and decoder can be specialized using different types of neural networks, such as GRU~\cite{bahdanau2014neural}, CNN~\cite{gehring2017convolutional}, and Transformer~\cite{vaswani2017attention}. 
    Considering the performance of existing SLT systems, we adopt the Transformer as well.
    
    With the goal of narrowing the domain gap between vision and text, we augment the original translation process with an iterative refinement step. Fig.~\ref{fig2} provides an overview of the proposed IP-SLT model, which consists of three stages, namely feature extraction, prototype initialization, and iterative prototype refinement. As with previous approaches, the initialization and iterative refinement module adopt the encoder-decoder architecture. Given the sign language video $\bm{X}=\{\bm{x}_t\}_{t=1}^{T_x}$ with $T_x$ frames, the feature extraction module embeds it into the spatial-temporal feature $\bm{F}=\{\bm{f}_t\}_{t=1}^{T_f}$. Next, the encoder $\mathcal{E}_1$ and decoder $\mathcal{D}_1$ are employed in the initialization module to derive the initial prototype $\bm{E}^0$ and initial translation $\bm{Y}^0$ from the visual feature $\bm{F}$. 
    
    Subsequently, the refinement module iteratively refines the previous prototype and generates the final translation $\bm{Y}^K=\{\bm{y}^K_t\}_{t=1}^{T_{y,K}}$ with $T_{y,K}$ words after total $K$ iterations.
    At the $k$-th iteration, the encoder $\mathcal{E}_2$ estimates the prototype $\bm{E}^k$ for the current step by augmenting the original visual feature $\bm{F}$ with the previous prototype $\bm{E}^{k-1}$. 
    Next, the decoder $\mathcal{D}_2$ predicts the corresponding translation $\bm{Y}^k$. 
    Finally, the translation sequence $\bm{Y}=\{\bm{Y}^0,\bm{Y}^1,\cdots,\bm{Y}^K\}$ is obtained according to the prototype sequence $\bm{E}=\{\bm{E}^0,\bm{E}^1,\cdots,\bm{E}^K\}$. For the IP-SLT optimization, by dividing the decoding part into $K+1$ branches, we add iterative distillation supervision from the final translation to the middle translations. Since the refinement process takes place in the encoder $\mathcal{E}_2$, in inference, the proposed IP-SLT can generate the translation directly from the $K$-th prototype which causes acceptable overhead.

    \subsection{Feature Extraction}~\label{Sec3.2}
    The feature extraction module embeds a series of video frames $\bm{X}\in\mathbb{R}^{T_x\times H\times W \times 3}$ with width $W$ and height $H$ into its visual feature $\bm{F}\in\mathbb{R}^{T_f\times C}$ with the dimension of feature $C$. Since its goal is to extract a distinguishable representation for SLT, we can draw on the visual backbone used in CSLR~\cite{cui2019deep, koller2016deep,koller2019weakly, cui2017recurrent} to extract the valid representation. 
    Generally, with sliding window size $w$ and stride size $s$, the sign video is split into $T_f=\left \lceil \frac{T_x}{s}  \right \rceil $ clips. By passing sign videos through it, the spatial-temporal embeddings $\bm{F}=\{\bm{f}_t\}_{t=1}^{T_f}$ are extracted as:
    \begin{equation}
    \begin{aligned}
    \label{equ:1}
    \{\bm{f}_t\}_{t=1}^{T_f}=Extractor(\{\bm{x}_t\}_{t=1}^{T_x}).
    \end{aligned}
    \end{equation}
    
    \subsection{Prototype Initialization} 
    After a visual feature $\bm{F}\in\mathbb{R}^{T_f\times C}$ is extracted by the feature extraction module, it is first fed into the initialization module. The initialization module consists of an encoder $\mathcal{E}_1$ and a decoder $\mathcal{D}_1$.

    The visual representation $\bm{F}$ is first fed into the encoder $\mathcal{E}_1$, and encoded into $T_f$ raw states $\bm{E}^0=\{\bm{e}^0_t\}_{t=1}^{T_f}\in \mathbb{R}^{T_f\times C}$. Then the decoder $\mathcal{D}_1$ reads the prototype $\bm{E}^0$ and produces the initial translation $\bm{Y}^0=\{\bm{y}^0_t\}_{t=1}^{T_{y,0}}$ according to the predicted logits $\bm{U}^0=\{\bm{u}^0_t\}_{t=1}^{T_{y,0}}$. 
    Specifically, it predicts the conditional probability of the translation sequence, which is formulated as: \vspace{-1mm}
    \begin{equation}
    \begin{aligned}
    \label{equ:2}
    p^{0-th}(\bm{Y}^0|\bm{X})=\prod_{t=1}^{T_{y,0}}p(\bm{y}^0_t|\bm{F},\bm{y}^0_{0:t-1}),
    \end{aligned}
    \end{equation}
    where $\bm{y}^0_{0:t-1}=\{\bm{y}^0_0, \bm{y}^0_1, \dots, \bm{y}^0_{t-1}\}$ denotes the previous output sub-sequence at the $t$-th step. The initial token $\bm{y}^0_0$ represents the beginning of a sentence.
    The predicted probability of each token in the translation is computed as:\vspace{-1mm}
    \begin{equation}
    \begin{aligned}
    \label{equ:3}
    p(\bm{y}^0_t|\bm{F},\bm{y}^0_{1:t-1})=&softmax(\bm{u}^0_t)_{\bm{y}^0_t}\\
    =&softmax(\bm{h}^0_t\cdot\bm{W})_{\bm{y}^0_t},
    \end{aligned}
    \end{equation}
    where $\bm{h}^0_t\in\mathbb{R}^{C}$ represents the output of the final layer at the $t$-th step, and $\bm{W}\in\mathbb{R}^{C\times \left |\mathcal{V}\right |}$ denotes a linear mapping to projects the hidden state $\bm{h}^0_t$ into the predicted logits over the target vocabulary $\mathcal{V}$. The probability is calculated by applying the $softmax(\cdot)$ function to the logits.
    Notably, our goal is to obtain a more accurate prototype for SLT, thus, the decoder $\mathcal{D}_1$ is only used in the training process to provide guidance for the initialization module.

    \subsection{Iterative Prototype Refinement}
    
    \begin{figure}
    \centering
    \includegraphics[width=63mm]{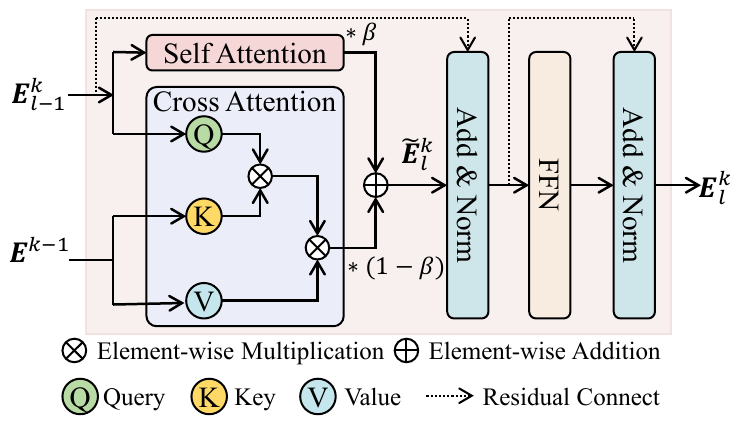}
    \caption{An illustration of the prototype update process in the $l$-th layer of the encoder $\mathcal{E}_2$ at the $k$-th iteration.
    }
    \label{fig3}
    \end{figure}
    
    Once the initial prototype $\bm{E}^0$ is obtained through the initialization module, we feed it together with the original visual representation $\bm{F}$ into the iterative refinement module. The module maintains a single prototype which is iteratively refined. In this way, the coarse semantic feature finally converges to a stable state where the prototype best fits the sign language semantics. The refinement module is divided into two sub-processes, \emph{i.e.}, iterative prototype aggregation in the encoder $\mathcal{E}_2$ and translation generation in the decoder $\mathcal{D}_2$.

    \smallskip
    \noindent \textbf{Iterative prototype aggregation}.
    To utilize the prototype as a reference, each layer of the encoder $\mathcal{E}_2$ attends over the maintained semantic features through the attention mechanism. 
    As shown in Fig.~\ref{fig3}, we illustrate the prototype aggregation process in the $l$-th layer of the encoder $\mathcal{E}_2$ at the $k$-th iteration. 
    The input of the refinement module consists of the original visual feature $\bm{F}$ and the $(k-1)$-th prototype $\bm{E}^{k-1}$. The attention mechanism $\text{attn}(\bm{q},\bm{K},\bm{V})$ is originally used in Transformer~\cite{vaswani2017attention}, which is formulated as:\vspace{-1mm}
    \begin{equation}
    \begin{aligned}
    \label{equ:3}
    &\text{attn}(\bm{q},\bm{K},\bm{V})=\sum_{i=1}^{\left | V \right | } \alpha_i\bm{W}_v\bm{v}_i, \\
    &\alpha_i=\text{softmax}((\bm{W}_q\bm{q})^T(\bm{W}_k\bm{k}_i)),
    \end{aligned}
    \end{equation}
    where $\bm{W}_q,\bm{W}_k$ and $\bm{W}_v$ are learnable parameters. To make better use of the $(k-1)$-th prototype, we consider the $(k-1)$-th prototype $\bm{E}^{k-1}=\{\bm{e}_t^{k-1}\}_{t=1}^{T_f}$ as key-value pair, and the original visual feature $\bm{F}=\{\bm{f}_t\}_{t=1}^{T_f}$ as the query. In this way, we inject the rich semantic information into the new prototype $\bm{E}^{k}=\{\bm{e}_t^{k}\}_{t=1}^{T_f}$.
    
    The encoder $\mathcal{E}_2$ is composed of $L_{e}$ identical layers, where $\bm{E}_l^{k}$ denotes the output of the $l$-th layer at the $k$-th iteration. 
    Similar to~\cite{zhu2020incorporating}, the hidden state is computed as:
    \begin{equation}
    \begin{aligned}
    \label{equ:4}
    \widetilde{\bm{E}}_l^{k}=&\beta \cdot \text{attn}_s(\bm{E}_{l-1}^{k},\bm{E}_{l-1}^{k},\bm{E}_{l-1}^{k})\\
    &+(1-\beta)\cdot \text{attn}_c(\bm{E}_{l-1}^{k},\bm{E}^{k-1},\bm{E}^{k-1}),
    \end{aligned}
    \end{equation}
    where $\text{attn}_s$ and $\text{attn}_c$ denote the self-attention sub-layer and cross-attention sub-layer used in Transformer~\cite{vaswani2017attention}, respectively.
    For the first layer of the encoder $\mathcal{E}_2$, $\bm{E}_{l-1}^{k}$ is equal to the visual feature $\bm{F}$.
    The output of the encoder $\mathcal{E}_2$ is $\bm{E}_{L_e}^{k}$ (\emph{i.e.}, $\bm{E}^{k}$).
    $\beta \in \left[0,1\right]$ is a hyperparameter that weights the importance of previous prototypes during training and inference. To further fuse and refine the prototype, it is linked with a fully connected sub-layer $\text{FFN}(\cdot)$ using the residual connection. The output of the $l$-th layer at the $k$-th iteration is formulated as:
    \begin{equation}
    \begin{aligned}
    \label{equ:5}
    \bm{E}_l^{k}=\text{LN}(\text{FFN}(\text{LN}(\widetilde{\bm{E}}_l^{k}+\bm{E}_{l-1}^{k}))+\text{LN}(\widetilde{\bm{E}}_l^{k}+\bm{E}_{l-1}^{k})),
    \end{aligned}
    \end{equation}
    where $\text{LN}(\cdot)$ is the layer normalization operation.

    \smallskip
    \noindent \textbf{Translation generation}.
    The decoder $\mathcal{D}_2$ iteratively takes the prototype $\bm{E}^k$ as input and generates the corresponding translation $\bm{Y}^k$. Take the $k$-th iteration as an example. 
    The decoder $\mathcal{D}_2$ predicts the conditional probability of translation $\bm{Y}^k=\{\bm{y}^k_t\}_{t=1}^{T_{y,k}}$ based on the predicted logits $\bm{U}^k=\{\bm{u}^k_t\}_{t=1}^{T_{y,k}}$, which is computed as:\vspace{-1mm}
    \begin{equation}
    \begin{aligned}
    \label{equ:6}
    p^{k-th}(\bm{Y}^k|\bm{X})=\prod_{t=1}^{T_{y,k}}p(\bm{y}^k_t|\bm{F},\bm{E}^{k-1}, \bm{y}^k_{0:t-1}),\\
    p(\bm{y}^k_t|\bm{F},\bm{E}^{k-1}, \bm{y}^k_{0:t-1})=softmax(\bm{u}^k_t)_{\bm{y}^k_t},
    \end{aligned}
    \end{equation}\vspace{-1mm}
    where $\bm{u}^k_t$ is the predicted logits of the decoder $\mathcal{D}_2$ at the $t$-th step.
    The decoder $\mathcal{D}_2$ iteratively takes the prototype given by the encoder $\mathcal{E}_2$ as input to generate the final translation $\bm{Y}^K$. With total $K$ iterations, the outputs of encoder $\mathcal{E}_1$ and $\mathcal{E}_2$ compose the prototype sequence $\bm{E}=\{\bm{E}^k\}_{k=0}^K$, where $\bm{E}^k=\{\bm{e}_t^k\}_{t=1}^{T_f}$ is the semantic feature at the $k$-th iteration. Accordingly, the decoder $\mathcal{D}_1$ and $\mathcal{D}_2$ generate $K+1$ translations $\bm{Y}=\{\bm{Y}^k\}_{k=0}^K$, where $\bm{Y}^k=\{\bm{y}_t^k\}_{t=1}^{T_{y,k}}$ is the translation at the $k$-th iteration during training. Note that after $K$ iterations, the decoder $\mathcal{D}_2$ obtains the converged prototype $\bm{E}^K$ and generates the translation only once in inference. 
    
    \subsection{Training Objective}
    We introduce two kinds of losses in the training period of the IP-SLT system.
    Firstly, the cross entropy loss is adopted to supervise the final generated sentence.
    Secondly, we put forward an iterative distillation loss. As the decoder $\mathcal{D}_1$ and $\mathcal{D}_2$ generate translation sequence $\bm{Y}=\{\bm{Y}^0,\bm{Y}^1,\dots,\bm{Y}^K\}$, we naturally divide the sequence $\bm{Y}$ into an initial prediction, $K-1$ intermediate predictions, and the final prediction according to the order of iteration. 
    Conceptually, the $K-1$ intermediate predictions are regarded as the student model and distill knowledge from the final prediction which is regarded as the teacher model. 

    \smallskip
    \noindent \textbf{Cross entropy loss}.
    As mentioned above, the IP-SLT generates the translation sequence based on the conditional probability provided by the decoder $\mathcal{D}_1$ and $\mathcal{D}_2$. The cross-entropy loss~\cite{vaswani2017attention} is computed with the ground truth from $N$ training samples and the outputs of the decoder $\mathcal{D}_1$ and $\mathcal{D}_2$. Its training objective in the proposed approach is to maximize the log-likelihood which is equal to minimizing the cross entropy loss formulated as:\vspace{-1mm}
    \begin{equation}
    \begin{aligned}
    \label{equ:8}
    L_{CE,k}=-log~p^{k-th}(\hat{\bm{Y}}|\bm{X}), 
    \end{aligned}
    \end{equation}
    where $\hat{\bm{Y}}$ denotes translation annotation.
    We apply this at the output of the initialization module and the $K$-th output of the refinement module.

    \smallskip
    \noindent \textbf{Iterative distillation loss}.
    Since the KL (Kullback-Leibler) divergence loss can affect the teacher's networks to the student's networks, we compute it between the $K-1$ intermediate predictions and the final prediction. As the final translation is based on the previous prototypes, we are able to get better translations by approximating more vital characterization capabilities of middle prototypes. The iterative distillation loss (IDL) is formulated  as:
    \begin{equation}
    \begin{aligned}
    \label{equ:9}
    L_{IDL}= \sum_{k=1}^{K-1}KL(\bm{U}^{k},\bm{U}^{K}),
    \end{aligned}
    \end{equation}
    where $\bm{U}^{k}$ and $\bm{U}^{K}$ are the predicted output of the decoder $\mathcal{D}_2$ at the $k$-th and $K$-th iteration, respectively. 
    By computing the Kullback-Leibler divergence between the $\bm{U}^{k}$ and $\bm{U}^{K}$, the IP-SLT is encouraged to approximate the performance of the final iteration.
    We apply this at the $K-1$ shallow outputs of the refinement module.
    
    Overall, the loss function of our IP-SLT is formulated as:
    \begin{equation}
    \begin{aligned}
    \label{equ:10}
    L=L_{CE,0}+L_{CE,K}+\lambda \cdot L_{IDL}.
    \end{aligned}
    \end{equation}
    Under the guidance of CE loss, we first train the initial generation module until convergence as a warm start, and then apply the loss function as Equ.~\eqref{equ:10} for optimizing the IP-SLT system in an end-to-end manner. 

    \subsection{Inference Strategy}
    Since the refinement process only involves in the encoder $\mathcal{E}_2$, the initialization and iterative refinement modules behave differently during training and inference. The decoder $\mathcal{D}_1$ of the initialization module is leveraged to provide guidance for the encoder $\mathcal{E}_1$ during the training process, while it is not required in inference. Similarly, the autoregressive decoder $\mathcal{D}_2$ of the iterative refinement module generates translations for each iteration during training, while it just decodes the final translation once in inference.

    In inference, given a sign language video to be translated, the feature extraction module first converts it into a visual feature. The encoder $\mathcal{E}_1$ of the initialization module transforms it into the raw prototype. Then, the encoder $\mathcal{E}_2$ of the refinement module further iteratively refines it by fusing it with the original visual feature. Finally, the decoder $\mathcal{D}_2$ generates the translation based on the final prototype.

\section{Experiments}

    \setlength{\tabcolsep}{4.3pt}
    \begin{table*}[!ht]
     \centering
     \resizebox{0.95\linewidth}{!}{
     \begin{tabular}{lcccccccccccc}
     \toprule
     \multirow{2}{*}{Methods} & \multirow{2}{*}{LLM} & \multicolumn{5}{c}{Dev} &\,\,\,& \multicolumn{5}{c}{Test} \\
         &\,\,\,& ROUGE &BLEU-1 &BLEU-2&BLEU-3&BLEU-4& &  ROUGE &BLEU-1 &BLEU-2&BLEU-3&BLEU-4 \\
         \midrule
         Joint-SLRT~\cite{camgoz2020sign} & \XSolidBrush          & -     & 47.26 & 34.40 & 27.05 & 22.38 & & -     & 46.61 & 33.73 & 26.19 & 21.32  \\
         PET~\cite{jin2022prior} & \XSolidBrush                   & -     & -     & -     & -     & -     & & 49.97 & 49.54 & 37.19 & 29.30 & 24.02  \\
         BN-TIN-Transf.~\cite{zhou2021improving} & \XSolidBrush       & 46.87 & 46.90 & 33.98 & 26.49 & 21.78 & & 46.98 & 47.57 & 34.64 & 26.78 & 21.68  \\
         STMC~\cite{zhou2021spatial} & \XSolidBrush               & 48.24 & 47.60 & 36.43 & 29.18 & 24.09 & & 46.65 & 46.98 & 36.09 & 28.70 & 23.65  \\
         IP-SLT & \XSolidBrush                                    & \textbf{54.43} & \textbf{54.10} & \textbf{41.56} & \textbf{33.66} & \textbf{28.22} & & \textbf{53.72} & \textbf{54.25} & \textbf{41.51} & \textbf{33.45} & \textbf{27.97}  \\
         \midrule
         MMTLB~\cite{chen2022simple} & \Checkmark               & 53.10 & 53.95 & 41.12 & 33.14 & 27.61 & & 52.65 & 53.97 & 41.75 & 33.84 & 28.39  \\
         TwoStream-SLT~\cite{chen2022two} & \Checkmark          & 54.08 & 54.32 & 41.99 & 34.15 & 28.66 & & 53.48 & 54.90 & 42.43 & 34.46 & 28.95  \\
         \bottomrule
     \end{tabular}}
     \vspace{1mm}
     \caption{Performance comparison of IP-SLT with methods for SLT on PHOENIX-2014T. 
     `LLM' denotes adopting pre-trained large language models.
     }
     \label{tab:slt-pho}
     \vspace{-1mm}
     \end{table*}
    
     \setlength{\tabcolsep}{4.3pt}
      \begin{table*}[!htb]
         \centering
         \resizebox{0.95\linewidth}{!}{
         \begin{tabular}{lcccccccccccc}
         \toprule
         \multirow{2}{*}{Methods} & \multirow{2}{*}{LLM} & \multicolumn{5}{c}{Dev} &\,\,\,& \multicolumn{5}{c}{Test} \\
         &\,\,\,& ROUGE &BLEU-1 &BLEU-2&BLEU-3&BLEU-4& &  ROUGE &BLEU-1 &BLEU-2&BLEU-3&BLEU-4 \\
             \midrule
             SL-Luong~\cite{camgoz2018neural} & \XSolidBrush          & 34.28 & 34.22 & 19.72 & 12.24 & 7.96  & & 34.54 & 34.16 & 19.57 & 11.84 & 7.56   \\
             Joint-SLRT~\cite{camgoz2020sign} & \XSolidBrush          & 37.06 & 37.47 & 24.67 & 16.86 & 11.88 & & 36.74 & 37.38 & 24.36 & 16.55 & 11.79  \\
             BN-TIN-Transf.~\cite{zhou2021improving} & \XSolidBrush   & 37.29 & 40.66 & 26.56 & 18.06 & 12.73 & & 37.67 & 40.74 & 26.96 & 18.48 & 13.19  \\
             IP-SLT & \XSolidBrush                                    & \textbf{44.33} & \textbf{45.26} & \textbf{31.77} & \textbf{22.87} & \textbf{16.74} & & \textbf{44.09} & \textbf{44.85} & \textbf{31.50} & \textbf{22.66} & \textbf{16.72} \\
             \midrule
             MMTLB~\cite{chen2022simple} & \Checkmark               & 53.38 & 53.81 & 40.84 & 31.29 & 24.42 & & 53.25 & 53.31 & 40.41 & 30.87 & 23.92  \\   
             TwoStream-SLT~\cite{chen2022two} & \Checkmark          & 55.10 & 55.21 & 42.31 & 32.71 & 25.76 & & 55.72 & 55.44 & 42.59 & 32.87 & 25.79  \\
             \bottomrule
         \end{tabular}}
         \vspace{1mm}
         \caption{Performance comparison of IP-SLT with methods for SLT on CSL-Daily.
         `LLM' denotes adopting pre-trained large language models.
         }
         \label{tab:slt-csl}
         \vspace{-1mm}
     \end{table*}

    \subsection{Experimental Setup}
    \smallskip
    \noindent \textbf{Dataset}.
    We evaluate our approach on two pubic sign language translation datasets, \emph{i.e.}, PHOENIX14T~\cite{camgoz2018neural} and CSL-Daily~\cite{zhou2021improving}. 
    Both datasets provide gloss-level and spoken-sentence-level annotations.
    The PHOENIX14T dataset~\cite{camgoz2018neural} is the first large-scale neural SLT dataset created with $9$ German sign language interpreters. 
    The dataset is split into a training set ($7,096$), a development set ($519$), and a test set ($642$).
    The CSL-Daily dataset~\cite{zhou2021improving} is a Chinese SLT dataset containing $20,654$ annotated sign language videos. We follow the previous experimental setting~\cite{zhou2021improving} and split it into the training, development, and test set.

    \smallskip
    \noindent \textbf{Evaluation metrics}.
    Following the work~\cite{zhou2021improving}, we quantitatively assess the quality of translations according to the BLEU-$N$~\cite{papineni2002bleu} and ROUGE~\cite{rouge2004package}.
    The BLEU-$N$ ($N$ ranges from $1$ to $4$) cares more about the accuracy of the predicted translation while the ROUGE cares more about the consistency of sentences. 
    For both evaluation metrics, a higher value indicates a better performance.
    
    \smallskip
    \noindent \textbf{Training settings}.
    We implement our approach on Pytorch~\cite{paszke2017automatic}. 
    The encoder and decoder of the initialization and refinement module consist of $3$ layers, respectively.
    The dimension of the feed-forward network is $2048$. 
    The visual feature of the STMC~\cite{zhou2021spatial} is $1024$-dimension while the visual feature of BN-TIN-Transf.~\cite{zhou2021improving} and VAC~\cite{min2021visual} are $512$-dimension.
    To alleviate over-fitting, we set dropout and attention head to $0.1$ and $8$, respectively. 
    The training optimizer is Adam~\cite{kingma2014adam}.
    During training, the learning rate is fixed to $5\times10^{-5}$. 
    To ensure the features provided by the previous prototype and original sign video are fully utilized, we apply the drop-net~\cite{zhu2020incorporating} during training, which effects Equ.~\eqref{equ:4}.
    During training, for any layer in the encoder $\mathcal{E}_2$, with probability $\beta$, the hidden state $\widetilde{\bm{E}}_l^{k}$ in Equ.~\eqref{equ:4} is the output of the self-attention sub-layer $\text{attn}_s$; with probability $1-\beta$, it is the output of the cross-attention sub-layer $\text{attn}_c$. 
    In inference, the hidden state $\widetilde{\bm{E}}_l^{k}$ is computed as Equ.~\eqref{equ:4}.

    \smallskip
    \noindent \textbf{Inference details}.
    In inference, we use the beam search strategy~\cite{wu2016google} to improve the decoding accuracy. 
    For the PHOENIX-2014T dataset and the CSL-Daily dataset, we set the beam search width and the length penalty to $3$ and $1.0$, respectively.
    To reduce the computational complexity, our IP-SLT just decodes once in inference for each input sign language video.

    \subsection{Comparison with State-of-the-Art Methods}
    We compare the proposed IP-SLT with the previous SLT systems on two public benchmarks, \emph{i.e.}, PHOENIX14T~\cite{camgoz2018neural} and CSL-Daily~\cite{zhou2021improving}, and the performance of our IP-SLT is shown in Tab.~\ref{tab:slt-pho} and Tab.~\ref{tab:slt-csl}, respectively. 
    For PHOENIX14T and CSL-Daily dataset, we adopt the STMC~\cite{zhou2021spatial} and BN-TIN-Transf.~\cite{zhou2021improving} as the baseline, respectively.
    Our IP-SLT follows the sign-to-text (S2T) paradigm, which directly transforms the sign language video into translation.
    Note MMTLB~\cite{chen2022simple} and TwoStream-SLT~\cite{chen2022two} adopt pre-trained large-scale language models that leverage more model parameters and extra resources than IP-SLT.

    By combining all proposed components together, our IP-SLT achieves substantial improvements against the baseline. 
    The IP-SLT achieves $28.22$ and $16.74$ BLEU-4 scores on the DEV set of PHOENIX14T and CSL-Daily, respectively. 
    The quantitative results demonstrate that our IP-SLT achieves promising performance improvements.
    Our IP-SLT delivers promising performance gains on DEV and test sets by leveraging the iterative refinement process. The results prove the advantage of aggregating the previous translation progress and the original visual representation, which distinguishes our IP-SLT from previous SLT systems.

    \subsection{Ablation Studies}
    In this section, we put forward several ablation studies on the DEV set of PHOENIX-2014T. Unless otherwise specified, we adopt the STMC~\cite{zhou2021spatial} as the baseline for the following experiments.

    \setlength{\tabcolsep}{2.4pt}
    \begin{table}[!t]
     \scriptsize
     \resizebox{1.0\linewidth}{!}{
     \begin{tabular}{lccccc}
     \toprule
         Setting                    & ROUGE &BLEU-1 &BLEU-2&BLEU-3&BLEU-4 \\
         \midrule
         Baseline                   & 50.10 & 50.21 & 37.12 & 29.41 & 24.31 \\
          ~+Refinement              & 51.22 & 51.04 & 38.40 & 30.61 & 25.39 \\
          ~+IDL                     & \textbf{54.43} & \textbf{54.10} & \textbf{41.56} & \textbf{33.66} & \textbf{28.22} \\
         \midrule
         $6$-$6$ Layers             & 50.49 & 50.58 & 37.57 & 29.64 & 24.53 \\
         \bottomrule
     \end{tabular}
     }
     \vspace{0.5mm}
     \caption{Effect of our proposed components. `Refinement' denotes applying the refinement process. `IDL' denotes applying the iterative distillation loss. 
     `$6$-$6$ layers' denotes enlarging the encoder and decoder of the baseline system from $3$ to $6$ layers.
     }
     \label{tab:A1}
     \vspace{-1mm}
     \end{table}

     \setlength{\tabcolsep}{1pt}
    \begin{table}[!t]
     \scriptsize
     \resizebox{1.0\linewidth}{!}{
     \begin{tabular}{lcccccc}
     \toprule
         Model                           & RM    & ROUGE & BLEU-1 & BLEU-2 & BLEU-3 & BLEU-4\\ 
         \midrule
         \multirow{2}{*}{STMC}           & w/o & 50.10 & 50.21 & 37.12 & 29.41 & 24.31 \\
                                         & w/   & \textbf{54.43} & \textbf{54.10} & \textbf{41.56} & \textbf{33.66} & \textbf{28.22} \\
        \midrule
        \multirow{2}{*}{BN-TIN-Transf.}  & w/o     & 47.41 & 47.99 & 34.94 & 27.33 & 22.35 \\
                                         &  w/   & \textbf{52.06} & \textbf{52.06} & \textbf{39.01} & \textbf{31.08} & \textbf{25.69} \\
         \midrule
         \multirow{2}{*}{VAC-Transf.}    & w/o     & 49.48 & 50.01 & 37.00 & 29.12 & 23.91 \\
                                         &  w/   & \textbf{53.68} & \textbf{53.60} & \textbf{41.28} & \textbf{33.47} & \textbf{28.07} \\
         \bottomrule
     \end{tabular}
     }
     \vspace{0.5mm}
     \caption{
     Generalization of IP-SLT.
     `RM' denotes leveraging the refinement process. 
     `w/' and `w/o' denote the baseline SLT system with and without a refinement process, respectively.
     }
     \label{tab:A3}
     \vspace{-1mm}
     \end{table}

    \smallskip
    \noindent \textbf{Impact of network architecture}.
    The main difference between our proposed method and the existing work is to leverage the previous information as an additional clue to enhance the current prototype. To evaluate the effectiveness of each component, we gradually add the refinement module and the iterative distillation loss to the baseline SLT system. Directly applying the refinement process to the baseline delivers a performance gain of $1.08$ BLEU-4. 
    We further apply the iterative distillation loss to improve the performance. The results suggest that adding distillation supervision can be helpful with a gain of $2.83$ BLEU-4. 
    Besides, to keep the number of parameters unchanged, we enlarge the depth from $3$ to $6$ layers and evaluate the performance. 
    Naively enlarging the model scale slightly improves the performance ($+0.22$ BLEU-4).
    The results are shown in Tab.~\ref{tab:A1}. 

     \setlength{\tabcolsep}{5.5pt}
    \begin{table}[!t]
     \scriptsize
     \resizebox{1.0\linewidth}{!}{
     \begin{tabular}{cccccc}
     \toprule
         $\lambda$                 & ROUGE &BLEU-1 &BLEU-2&BLEU-3&BLEU-4 \\
         \midrule
         0                         & 51.30 & 50.32 & 38.01 & 30.29 & 25.14 \\
         5                         & 53.07 & 52.31 & 40.08 & 32.29 & 27.02 \\
         10                        & 53.71 & 53.51 & 40.92 & 32.94 & 27.47 \\
         15                        & \textbf{54.43} & 54.10 & \textbf{41.56} & \textbf{33.66} & \textbf{28.22} \\
         20                        & 54.42 & \textbf{54.16} & 41.51 & 33.44 & 27.87 \\
         \bottomrule
     \end{tabular}
     }
     \vspace{0.5mm}
     \caption{The weight $\lambda$ of iterative distillation loss to CE loss.}
     \label{tab:A4}
     \vspace{-1mm}
     \end{table}

    \smallskip
    \noindent \textbf{Generalization of the IP-SLT}. 
    We conduct three sets of experiments by changing the visual backbone to evaluate the generalization of the proposed IP-SLT approach in Tab.~\ref{tab:A3}. Specifically, the BN-TIN-Transf.~\cite{zhou2021improving} uses a basic CNN network to get the dense representation of sign video. The STMC~\cite{zhou2021spatial} extracts and aligns the key parts of sign language expression to achieve better performance. The VAC~\cite{min2021visual} proposes two auxiliary supervision methods to enhance the feature extraction module.
    VAC-Transf. replaces the feature extractor in BN-TIN-Transf.~\cite{zhou2021improving} with the visual backbone of VAC~\cite{min2021visual}.
    Applying our proposed IP-SLT methods to the BN-TIN-Transf., VAC-Transf. and STMC, we achieve $25.69$, $28.07$, and $28.22$ BLEU-4 scores on the DEV set, surpassing the baselines by $3.34$, $4.16$, $3.91$, respectively. Using a high-quality visual backbone delivers further quality gains.
 
    \setlength{\tabcolsep}{5.0pt}
    \begin{table}[!t]
     \scriptsize
     \resizebox{1.0\linewidth}{!}{
     \begin{tabular}{cccccc}
     \toprule
         $\beta$                      & ROUGE &BLEU-1 &BLEU-2&BLEU-3&BLEU-4 \\
         \midrule
         0.0                           & 50.10 & 50.21 & 37.12 & 29.41 & 24.31 \\
         0.2                           & 54.35 & 53.81 & 41.40 & 33.52 & 28.02 \\
         0.4                           & 54.39 & 54.02 & 41.24 & 33.19 & 27.70 \\
         0.5                           & \textbf{54.43} & \textbf{54.10} & \textbf{41.56} & \textbf{33.66} & \textbf{28.22} \\
         0.6                           & 53.96 & 53.63 & 40.94 & 33.10 & 27.72 \\
         0.8                           & 54.31 & 53.39 & 40.92 & 32.98 & 27.53 \\
         \midrule
         -                             & 50.97 & 50.67 & 38.24 & 30.52 & 25.41 \\
         \bottomrule
     \end{tabular}
     }
     \vspace{0.5mm}
     \caption{The weight $\beta$ of the original visual feature to the previous prototype. `0.0' denotes the assessment of the baseline. `-' denotes setting $\beta$ as $0.5$ without using the drop-net~\cite{zhu2020incorporating}.}
     \label{tab:A7}
     \vspace{-1mm}
     \end{table}

     \setlength{\tabcolsep}{2.7pt}
    \begin{table}[!t]
     \scriptsize
     \resizebox{1.0\linewidth}{!}{
     \begin{tabular}{lccccc}
     \toprule
         Setting                        & ROUGE &BLEU-1 &BLEU-2&BLEU-3&BLEU-4 \\
         \midrule
         None                           & 50.10 & 50.21 & 37.12 & 29.41 & 24.31 \\
         Con-input                      & 54.37 & 53.83 & 41.10 & 32.97 & 27.39 \\
         Con-feature                    & 53.97 & 52.76 & 40.56 & 32.81 & 27.47 \\
         Add-feature                    & \textbf{54.43} & \textbf{54.10} & \textbf{41.56} & \textbf{33.66} & \textbf{28.22} \\
         \bottomrule
     \end{tabular}
     }
     \vspace{0.5mm}
     \caption{Effect of the different refinement methods. `Con-input' denotes directly concatenating the original visual feature and previous prototype as input. `Con-feature' denotes concatenating the original feature and the previous feature given by the cross-attention mechanism in each layer. On top of that, `Add-feature' denotes changing it to an addition operation.}
     \label{tab:A6}
     \vspace{-1mm}
     \end{table}

    \smallskip
    \noindent \textbf{Impact of $\lambda$}.
    In our experiments, the weight $\lambda$ of iterative distillation loss is set to $15$. It is a hyper-parameter that is designed to balance the effect of cross-entropy loss and the iterative distillation loss.
    We conduct experiments by varying the weight $\lambda$.
    Tab.~\ref{tab:A4} shows that our IP-SLT achieves the best performance when the weight $\lambda$ is set to$15$.

    \smallskip
    \noindent \textbf{Impact of $\beta$}.
    In the above experiments, the weight $\beta$ is fixed to $0.5$. The weight $\beta$ represents the importance of the previous prototype compared with the original visual feature. As a hyper-parameter of our proposed methods, the weight $\beta$ is examined with a set of different values in Tab.~\ref{tab:A7}.
    When the weight is $0.5$, the performance is the highest one.
    This indicates that the previous prototype to SLT is as important as the original visual feature. 
    Besides, to fully use the previous prototype, the drop-net is required.

    \smallskip
    \noindent \textbf{Impact of refinement method}.
    We also examine a set of refinement methods for IP-SLT considering the fusion mechanism as a key part of our proposed method in Tab.~\ref{tab:A6}. Directly concatenating the original visual feature and previous prototype in the time dimension improves the performance of SLT from $24.31$ to $27.39$. 
    We further add the cross-attention mechanism to explicitly leverage the useful information from the previous prototype. And then, we concatenate the representation from the original visual feature and the representation from the previous prototype in feature dimension, which delivers performance gains of $3.16$ BLEU-4. On top of that, changing the concatenating process to an element-wise addition operation achieves a further quality gain of $3.91$. 

    \smallskip
    \noindent \textbf{Impact of iteration number $K$}.
    The iteration number $K$ is an important hyper-parameter and is fixed to $3$ in the previous experiments. We conduct experiments with different iteration numbers to explore the effect of iteration number $K$. Tab.~\ref{tab:A5} shows that there is a culmination in the performance at iteration $3$. Before this culmination, the performance improvement of our proposed method increase fast. When the iteration number is bigger than $3$, the performance of our method is slightly weakened (-$0.21$ BLEU-4).

    \setlength{\tabcolsep}{5.5pt}
    \begin{table}[!t]
     \scriptsize
     \resizebox{1.0\linewidth}{!}{
     \begin{tabular}{cccccc}
     \toprule
         $K$                       & ROUGE &BLEU-1 &BLEU-2&BLEU-3&BLEU-4\\
         \midrule
         0                         & 50.10 & 50.21 & 37.12 & 29.41 & 24.31 \\
         1                         & 51.22 & 51.04 & 38.40 & 30.61 & 25.39 \\
         2                         & 53.73 & 53.39 & 40.76 & 32.98 & 27.66 \\
         3                         & 54.43 & \textbf{54.10} & \textbf{41.56} & \textbf{33.66} & \textbf{28.22} \\
         4                         & \textbf{54.63} & 53.91 & 41.40 & 33.53 & 28.01 \\
         \bottomrule
     \end{tabular}
     }
     \vspace{0.5mm}
     \caption{
     Effect of the iteration number $K$ in the iterative refinement module.
     }
     \label{tab:A5}
     \vspace{-1mm}
     \end{table}

     \setlength{\tabcolsep}{1.5pt}
    \begin{table}[!t]
     \scriptsize
     \resizebox{1.0\linewidth}{!}{
     \begin{tabular}{lcccccc}
     \toprule
         Model                       & SW   & I-P(M) & T-P(M) & FLOPs(B) & ROUGE & BLEU-4\\ 
         \midrule
         \multirow{3}{*}{STMC}                        & -     & 92.6  & 92.6  & 28.7 & 50.10 & 24.31 \\
                                     &  w/   & 128.7 & 172.6 & 32.6 & 54.43 & 28.22 \\
                                     &  w/o  & 238.6 & 357.9 & 32.6 & \textbf{54.81} & \textbf{28.48} \\
         \midrule
         \multirow{3}{*}{BN-TIN-Transf.}              & -     & 28.3  & 28.3  & 9.6 & 47.41 & 22.35 \\
                                     &  w/   & 37.7  & 53.4  & 10.6 & 52.06 & 25.69 \\
                                     &  w/o  & 66.1  & 110.0 & 10.6 & \textbf{52.24} & \textbf{25.96} \\
         \midrule
         \multirow{3}{*}{VAC-Transf.}                 & -     & 28.3  & 28.3  & 9.6 & 49.48 & 23.91 \\
                                     &  w/   & 37.7  & 53.4  & 10.6 & 53.68 & 28.07 \\
                                     &  w/o  & 66.1  & 110.0 & 10.6 & \textbf{54.82} & \textbf{28.27} \\
         \bottomrule
     \end{tabular}
     }
     \vspace{0.5mm}
     \caption{
     Comparing the baseline and non-shared weight prototype refinement method with IP-SLT.
     `SW' denotes the sharing weight across all iterations.
     `-' denotes the baseline SLT system.
     `w/' and `w/o' denote the refinement module in different iterations with and without shared parameters, respectively.
     `I-P' and `T-P' denote the parameter amount calculated in inference and during training, respectively. }
     \label{tab:A2}
     \vspace{-1mm}
     \end{table}
     
    \smallskip
    \noindent \textbf{Impact of sharing weights between different iterations}.
    In Tab.~\ref{tab:A2}, we examine the storage and translation quality of our proposed IP-SLT with different baselines \emph{i.e.}, STMC, BN-TIN-Transf, and VAC-Transf. In order to ensure the efficiency of the model, our refinement module shares weight across all iterations. 
    Since the feature extraction module of each group is identical, the parameters of the feature extraction module are not included in the calculation of the storage of different models.
    As in inference, several parts can be removed from IP-SLT without changing the performance, we report the number of parameters in both training and inference, respectively. The results indicate that our iterative process causes acceptable overhead with remarkable performance improvements. We further conduct experiments in which the parameters of each iteration are independent. The results demonstrate that the proposed parameter-shared method achieves near performance gain ($28.22$ v.s $28.48$ BLEU-4, $25.69$ v.s $25.96$ BLEU-4, and $28.07$ v.s $28.27$ BLEU-4) with non-shared one but leveraging much less parameter number.

    \smallskip
    \noindent \textbf{Computation comparison with baseline}. 
    The FLOPs is a key factor of computation efficiency. We conduct experiments to compare the computation of the proposed IP-SLT method with different baselines. 
    Similarly, we exclude the FLOPs of the feature extraction module and report relevant computational costs in inference.
    Tab.~\ref{tab:A2} shows that leveraging the iterative refinement process can cause acceptable computation costs while achieving promising performance.

    \smallskip
    \noindent \textbf{Case Study}.
    To provide a more intuitionistic view of our proposed method, we list some translation samples of the proposed IP-SLT method in Tab.~\ref{tab:A8}. We observe that based on the previous prototype and original visual representation, the IP-SLT can generate more accurate and fluent sentences. 
    
    \begin{table}[!t]
    \centering
    \scriptsize
    \setlength{\tabcolsep}{3.5pt}
    \resizebox{1.0\linewidth}{!}{
    \begin{tabular}{ll}
    \toprule
         \multicolumn{1}{l}{Type} & \multicolumn{1}{l}{Text}\\
         \midrule
         GT                    & ich wünsche ihnen einen schönen abend und machen sie es gut. \\
         Baseline              & jetzt wünsche ich ihnen noch einen schönen abend. \\
         Our                   & \textbf{ihnen einen schönen abend und machen sie es gut.} \\
        \midrule
         GT                    & in der neuen woche wird es milder aber es bleibt wechselhaft. \\
         Baseline              & dann wird es wieder milder. \\
         Our                   & \textbf{in der neuen woche wird es dann wieder milder.} \\ 
         \bottomrule
     \end{tabular}}
     \vspace{0.5mm}
     \caption{
     Qualitative evaluation.
     `GT' denotes the spoken language translation annotation. `Baseline' and `Our' denote the translation result of baseline and our IP-SLT, respectively.}
     \label{tab:A8}
     \vspace{-1mm}
     \end{table}

\section{Conclusion}
In this work, we propose a new framework IP-SLT which introduces the iterative refinement into a conventional SLT system. With the goal to polish the semantic representation by leveraging the previous results, we present IP-SLT to support the iterative refinement process. The proposed method is differentiable and optimized in an end-to-end manner to achieve its best performance.
On top of it, we put forward the iterative distillation loss to further improve the translation quality by leveraging the sequential dependence between the outputs of each iteration.
In inference, the autoregressive decoding process is required once to generate the translation based on the final prototype, applying IP-SLT does not significantly affect efficiency.
The experimental results demonstrate the effectiveness of the IP-SLT.

\smallskip \noindent \textbf{Acknowledgments:} This work was supported by NSFC under Contract U20A20183 and 62021001. It was also supported by the GPU cluster built by MCC Lab of Information Science and Technology Institution, USTC, and the Supercomputing Center of the USTC.

{\small
\bibliographystyle{ieee_fullname}
\bibliography{egbib}
}

\end{document}